\documentclass[runningheads]{llncs}
\usepackage[T1]{fontenc}
\usepackage{graphicx}
\usepackage{booktabs} % for professional tables
\usepackage{array,multirow,graphicx}
\usepackage{color, colortbl}
\usepackage[dvipsnames]{xcolor}
\usepackage{caption}
\usepackage{tikz}
\usepackage{tablefootnote}
\newcommand*\circled[1]{\tikz[baseline=(char.base)]{
            \node[shape=circle,draw,inner sep=0.1pt] (char) {#1};}}

\usepackage{soul}

\newcommand{\hlc}[2][yellow]{{%
    \colorlet{foo}{#1}%
    \sethlcolor{foo}\hl{#2}}%
}
\usepackage{cite}
\usepackage[colorlinks=true, urlcolor=blue, linkcolor=red]{hyperref}
\begin{document}
\title{Are We Ready for Out-of-Distribution\\Detection in Digital Pathology?}

\titlerunning{Are We Ready for OOD Detection in DP}
% \author{Anonymous}
\author{Ji-Hun Oh\inst{1}\and
Kianoush Falahkheirkhah\inst{1}\and
Rohit Bhargava\inst{1, 2}\\
}
%
% \authorrunning{Anonymous}
\authorrunning{JH Oh et al.}

\institute{
\textsuperscript{1} University of Illinois Urbana-Champaign, Urbana, IL, USA\\
\textsuperscript{2} CZ Biohub Chicago, LLC, Chicago, IL, USA\\
{\tt\small \{jihunoh2, kf4, rxb\}@illinois.edu}\\
}
\maketitle              % typeset the header of the contribution

\begin{abstract}
The detection of semantic and covariate out-of-distribution (OOD) examples is a critical yet overlooked challenge in digital pathology (DP). Recently, substantial insight and methods on OOD detection were presented by the ML community, but how do they fare in DP applications? To this end, we establish a benchmark study, our highlights being: 1) the adoption of proper evaluation protocols, 2) the comparison of diverse detectors in both a single and multi-model setting, and 3) the exploration into advanced ML settings like transfer learning (ImageNet \textit{vs.} DP pre-training) and choice of architecture (CNNs \textit{vs.} transformers). Through our comprehensive experiments, we contribute new insights and guidelines, paving the way for future research and discussion. We continuously update our code at \url{https://github.com/jihunoh2/OODD4DP}.

\end{abstract}

\section{Introduction}
The fickleness and fragility of deep neural networks (DNNs) makes them prone to overconfident but erroneous predictions, particularly under distribution shifts. In high-stake domains like digital pathology (DP), a subsequent misdiagnosis can be catastrophic, thus far hindering real-world DNN deployments. To facilitate trustworthy AI practices in DP, it is pivotal for DNNs to communicate ``\textit{I don't know}'' when unsure of its own prediction, allowing clinicians to intervene.

\textbf{Background.} Often pursued as an \textit{out-of-distribution (OOD) detection} problem, let us denote $f: \mathcal{X} \rightarrow \mathcal{Y}$ the classifier model, the focus of our article. Letting $P(\mathcal{X}_{ID},\mathcal{Y}_{ID}) \subseteq \mathcal{X}\times\mathcal{Y}$ the joint in-distribution (ID) defined by the training set, a sample is OOD if $(x, y)\notin P(\mathcal{X}_{ID},\mathcal{Y}_{ID})$. In literature, OOD is categorized into either \textit{semantic} or \textit{covariate}; semantic OOD arise from label-altering shifts $y \notin P(Y|X)$, whereas covariate OOD preserves the ID labels $y \in P(Y|X)$ but are modified by the image space $x \notin P(X)$. Note, $y \notin P(Y|X)$ generally entails $x \notin P(X)$, and thus, semantic OOD is shifted by the image space too. Since the model cannot inherently handle semantic OOD, we seek to flag them using some scoring function expressing OOD extent; examples include (\textit{but is not restricted to}) the softmax uncertainty. However, in the case of covariate OOD, unconditional detection is undesirable as it conflicts with \textit{generalization},\footnote{This is because covariate OODs share the ID labels, resulting in robust models capable of generalizing across such out-of-domains.} limiting its open-world application. To avoid such a dilemma, the emerging consensus in the ML community is to detect just its failures (\textit{i.e.}, misclassified instances) \cite{jaeger2022call, guerin2023out}. From here onwards, we refer to semantic and misclassified covariate OOD detection as S-OODD and MC-OODD, respectively.

In the field of DP, S-OODD is required foremost because models are conceived in a closed-set environment. For instance, many breast datasets do not include borderline atypical lesion or rarer carcinoma subtypes due to its rarity and/or annotating difficulties. Likewise, the need for MC-OODD is inevitable because DNNs are prone to overfit or memorize, even finding shortcuts by picking up spurious correlations (\textit{e.g.}, site-specific attributes coming from the staining and scanning procedures) in lieu of pathological generalization.

\textbf{Related work.} Few work have studied detection tasks in DP \cite{thagaard2020can, cao2020benchmark, linmans2023predictive}. Ref. \cite{thagaard2020can} inspected MC-OODD (across different organ \& hospital) and S-OODD (head/neck SCC) over the Camelyon\footnote{\url{https://camelyon17.grand-challenge.org/Home/}} dataset. Ref. \cite{cao2020benchmark} benchmarked various OOD detectors in PatchCamelyon, designating external datasets as OOD. Recently, \cite{linmans2023predictive} investigated the detection of diffuse large B-cell lymphoma \textit{w.r.t.} Camelyon, and prostate images containing colorectal sections \textit{w.r.t.} colon-free prostate biopsies. We however recognize several deficiencies in these work.

\textit{i) Misleading practice.} Not all detection objectives therein conform to the above-mentioned consensus. For instance, let us consider the example of detecting colorectal invasion in prostate biopsies \cite{linmans2023predictive}, which arises due to their anatomical proximity. Such a shift from co-occurrence does not necessarily induce a label change and is closer to a covariate OOD, thus, calling for MC-OODD. Note, it is possible for certain cases to potentially alter the patient's treatment/prognosis and become semantic OOD; in these scenarios, their detection makes sense, but such a distinction is not made in the above work. In addition, the goal of MC-OODD is to quantify how separable misclassifications are from its correct counterpart, \textit{regardless of their ratios}. However, using AUROC or AUPR like in \cite{thagaard2020can} yields a systemic bias as these metrics are sensitive to the model's accuracy \cite{malinin2019ensemble, pinto2022impartial}, preventing a fair study when comparing factors affecting OOD detection performance {but also simultaneously} the model's base performance \textit{e.g.}, choice of DNN architecture. 

\textit{ii) Limited OOD detectors.} Many works \cite{thagaard2020can, linmans2023predictive} adopt multi-model uncertainty quantification (UQ) to score OOD-ness, \textit{e.g.}, ensembles or approximate Bayes. However, these uncertainty measures like Shannon Entropy and Mutual Information can be falsely low far away from the ID data \cite{henning2021bayesian,ulmer2021know}. While \cite{cao2020benchmark} explored diverse detectors beyond those UQ-based, all are from 2020 or prior. More recent SOTA methods such as ViM \cite{wang2022vim} have yet to be explored in DP. 

\textit{iii) Easy or non-public datasets.} The datasets used are simple such as the binary classification task in Camelyon. In addition, some ``OODs'' therein are \textit{far} (or \textit{very-far}) \textit{w.r.t.} ID, making them easy to discern, \textit{e.g.}, prostate \textit{vs.} colon \textit{vs.} breast lymph nodes \cite{linmans2023predictive}, and H\&E \textit{vs.} other dyes \cite{cao2020benchmark}. Oftentimes, the work is also not reproducible because internal datasets were used \cite{thagaard2020can, linmans2023predictive}.

\textit{iv) Limited depth.} They overlook crucial factors known to influence robustness and OOD behavior like pre-training \cite{hendrycks2019usingpretrain} and DNN architecture \cite{pinto2022impartial}.

\begin{figure}[t]
\centering
\includegraphics[width=0.5\textwidth]{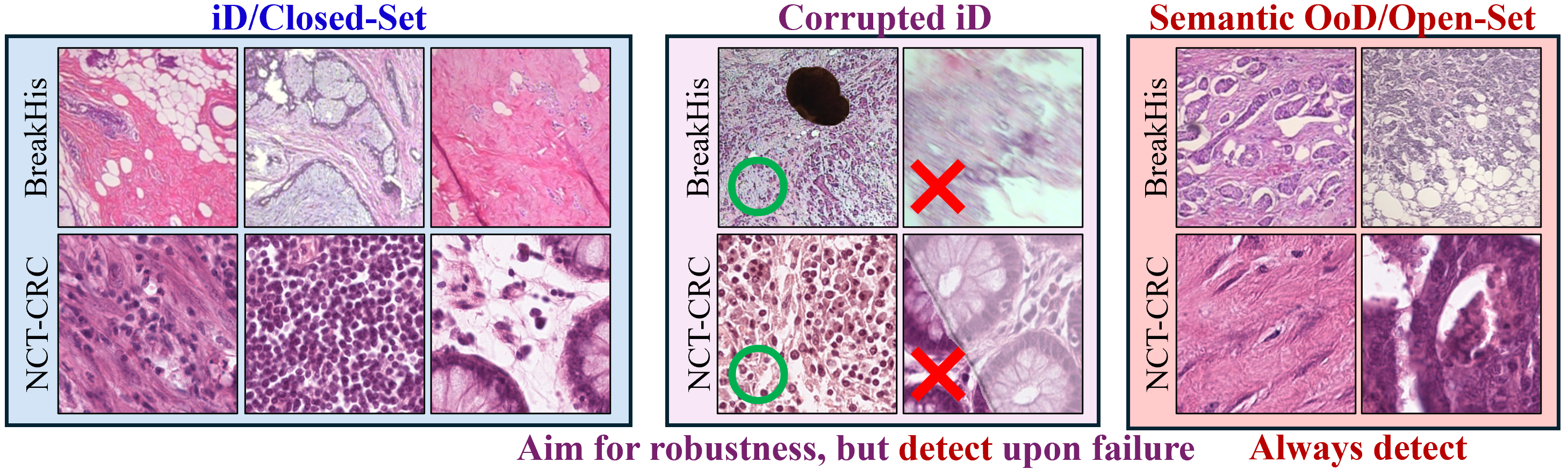}
\caption{\textbf{Our experiments.} We aim to i) generalize over corrupted (covariate-shifted) ID or detect its misclassification, and ii) always detect semantic OOD.} \label{fig1}
\end{figure}

\textbf{Contributions.} Acknowledging these shortcomings, we herein present an enhanced benchmark study. Concretely, we contribute in the following ways.

\textit{i) Proper protocols.} Adopting public datasets, we simulate an Open Set Recognition (OSR) setting by excluding a small fraction of the class during training and afterwards perform S-OODD \textit{w.r.t.} the held-out classes, permitting S-OODD evaluation in an {indisputable} way. As for MC-OODD, we apply common DP corruptions in \cite{zhang2022benchmarking} to the ID test set and report the Prediction Rejection Ratio (PRR) \cite{malinin2019ensemble}, a metric {agnostic to the model accuracy}. Our experiments (illustrated in Fig. \ref{fig1}) are free from any of the above-mentioned bias/malpractice or shortcomings, providing thus the most objective and reproducible benchmark to date on OOD detection in the context of DP.

\textit{ii) Wider scope.} In addition to UQ scores, we include a variety of recent frequentist (\textit{i.e.}, single-model) detection methods from top-tier ML conferences. We also investigate the impact of transfer learning (TL), specifically when adopting pre-trained DNN weights conceived from natural and DP images. Last but not least, we comparatively explore different architectures, namely fully-convnets (CNNs) \textit{vs.} transformers, the latter having seen a surge in popularity lately.

\textit{iii) Novel insights.} Through our extensive experiments, we answer questions like: Is there a detector best suited for S-OODD and/or MC-OODD? Does ensemble-based UQ (widely regarded the gold-standard) really perform best? Should we always pre-train over DP data? Are transformers better than CNNs like recent studies suggest \cite{bai2021transformers}? Whereas these are questions are popular subjects in the ML OOD detection literature, they have yet to be recognized and studied in the broader DP community as OOD detection has just recently started to gain traction. Our findings serve as a guide for practitioners and open up research questions and discussions for the future.

\section{Benchmark Setup}
Our overall pipeline involves training a classifier using an ID train set, and afterwards evaluating S-OODD and MC-OOD by respectively performing the binary classification of \textit{semantic OOD \textit{vs.} ID test set} and \textit{correctly classified covariate OOD vs. misclassified covariate OOD} using some OOD detection method. 

\begin{table*}[b]
\small
\centering
\caption{\textbf{OSR configs.} I and \hlc[red!50]{O} denotes ID (closed-set) and semantic OOD (open-set), respectively. The four/three-fold OSR split is arbitrary, aimed to reduce bias in our analysis. \tiny{\textsuperscript{\dag}B, Benign; A, Adenosis; F, Fibroadenoma; PT, Phyllodes Tumor; TA, Tubular Adenoma; M, Malignant; DC, Ductal Carcinoma (C); LC, Lobular C; MC, Mucinous C; PC, Papillary C. \textsuperscript{\ddag}ADI, adipose; BACK, background; DEB, debris; LYM, lymphocytes; MUC, mucus; MUS, muscle; NORM, normal mucosa; STR, C-associated stroma; TUM, adenocarcinoma epithelium.}}
\resizebox{0.8\textwidth}{!}{%
\begin{tabular}{c|cccccccc|ccccccccc}
\toprule
& & \multicolumn{6}{c}{BreakHis\textsuperscript{\dag} \cite{spanhol2015dataset}} & & & \multicolumn{7}{c}{NCT-CRC\textsuperscript{\ddag} \cite{kather2019predicting}} & \\ 
OSR & B-A & B-F & B-TA & B-PT & M-DC & M-LC & M-MC & M-PC & ADI & BACK & DEB & LYM & MUC & MUS & NORM & STR & TUM \\
\midrule
no. 1 & \cellcolor{red!50}O & I & I & I & \cellcolor{red!50}O & I & I & I & \cellcolor{red!50}O & I & I & \cellcolor{red!50}O & I & I & \cellcolor{red!50}O & I & I \\
no. 2 & I & \cellcolor{red!50}O & I & I & I & \cellcolor{red!50}O & I & I & I & \cellcolor{red!50}O & I & I & \cellcolor{red!50}O & I & I & \cellcolor{red!50}O & I \\
no. 3 & I & I & \cellcolor{red!50}O & I & I & I & \cellcolor{red!50}O & I & I & I & \cellcolor{red!50}O & I & I & \cellcolor{red!50}O & I & I & \cellcolor{red!50}O \\
no. 4 & I & I & I & \cellcolor{red!50}O & I & I & I & \cellcolor{red!50}O & & & & & & & & &  \\
\bottomrule
\end{tabular}}
\label{tab1}
\end{table*}

\textbf{Datasets.} We consider two public H\&E image datasets: \textbf{BreakHis} \cite{spanhol2015dataset}, consisting of eight breast carcinoma subtypes, and \textbf{NCT-CRC} \cite{kather2019predicting}, including nine colorectal tissue types. We simulate OSR splits in Tab. \ref{tab1}. Overall, the semantic OOD includes both \textit{near (hard)} and \textit{far (easy)} cases. For instance, the fine-grained MUS in NCT-CRC is hard to detect \textit{w.r.t.} STR (and vice versa) due to their shared fibrous structure. In contrast, we anticipate ADI, BACK (NCT-CRC) or M-LC (BreakHis) to be much easier given their drastically different appearances. For BreakHis, we perform a 4:1 train-test split among the ID samples, while using the CRC-VAL as the test set for NCT-CRC. To generate covariate OOD, we apply the DP corruptions from \cite{zhang2022benchmarking} with a severity level of 3, which includes several types of digitization, blur, color, and artifact.

\textbf{Training details.} All classifiers are trained using the focal loss. We crop images at 224×224 and apply random flips and transpose. We allocate a small hold-out subset for validation purposes, evaluating its accuracy at every 25 iterations. Using a mini-batch size of 32, we initialize the Adam learning rate to 1e-4 and downstep by 0.3 upon reaching the [2\textsuperscript{nd}, 4\textsuperscript{th}, 5\textsuperscript{th}, 6\textsuperscript{th}] times the validation set's performance fail to improve, terminating the training afterwards.

\textbf{ML configs.} It is well-established that TL by initializing from pre-trained weights of another task improves robustness in downstream applications \cite{hendrycks2019usingpretrain}. The de-facto protocol in computer vision is to fine-tune from ImageNet-1K (IN1K) weights. However, it is common belief that TL is more beneficial when the pre- and fine-tuning sources are similar in domain to capitalize feature reuse. Multiple recent works have thus proposed general-purpose DP models for TL, developed via non-supervised learning over massive amounts of unlabeled data \cite{kang2023benchmarking, radford2021learning, zhang2023large, ikezogwo2024quilt}. We delve into both TL options across two DNN backbones: \textbf{(C1) ResNet-50:} We compare training from scratch (\textit{i.e.}, no TL; de-novo) and TL from IN1K and three DP-specific models in \cite{kang2023benchmarking}, each embodying a distinct self-supervised learning (SSL) method trained with 19M H\&E patches extracted from The Cancer Genome Atlas (TCGA): MoCo v2 \cite{chen2020improved}, SwAV \cite{caron2020unsupervised}, and BT \cite{zbontar2021barlow}. \textbf{(C2) ViT-B/16} \cite{dosovitskiy2020image}: We compare de-novo, IN1K, and three foundational large vision-language models (LVLMs): CLIP \cite{radford2021learning}, BiomedCLIP \cite{zhang2023large}, and QuiltNet \cite{ikezogwo2024quilt}, in order of decreasing domain gap \textit{w.r.t.} our downstream datasets. CLIP is trained with 300M natural image-caption pairs, BiomedCLIP incorporates 15M pairs from PubMed but does not include H\&E, whereas QuiltNet utilizes 1M pairs of H\&E-text pairs scraped from MedTwitter and YouTube. We take the IN1K and CLIP pre-trained weights from the public timm-repository.\footnote{Timm names: \texttt{resnet50.a1\_in1k}, \texttt{vit\_base\_patch16\_224.augreg2\_in21k\_ft\_in1k}, \texttt{vit\_base\_patch16\_clip\_224.openai}.}

DNN architecture is also known to influence OOD behavior. We further inspect \textbf{(C3) different architectures}, notably CNNs (ResNet, ConvNeXt \cite{liu2022convnet}) \textit{vs.} transformers (ViT, Swin \cite{liu2021swin}). For a fair comparison, it is crucial to employ the same TL recipe to minimize its influence, but is not possible with existing pre-trained checkpoints which are limited to just a single architecture backbone. Hence, we pre-train from scratch using the 32 class pan-cancer TCGA dataset with patch-level annotations from \cite{komura2022universal}, adopting a training recipe similar to the above. Note, unlike (C1) \& (C2), this pre-training is \textit{fully-supervised}, developed in a much \textit{light-weight} manner with $\sim$272K patches, trainable in a single commercial GPU in less than a few hours, presenting hence a unique perspective on DP-specific TL. We dub our checkpoints as SIAYN (short for \textit{Supervision Is All You Need}) and make them public along with our code.

\textbf{Detectors.} In recent years, a plethora of OOD detection methods were proposed. We focus on \textbf{post-hoc/frequentist} (\textit{i.e.}, single-model inference-time) methods which are competitive and (almost) cost-free, summarized in Tab. \ref{tab2}: Max Softmax Probability (MSP) \cite{hendrycks2016baseline}, Mahalanobis distance (Maha) \cite{lee2018simple}, ReAct followed by Energy (R+E) \cite{sun2021react}, GradNorm (GrN) \cite{huang2021importance}, Max-Logit Score (MLS) \cite{hendrycks2022scaling}, KL-Matching (KLM) \cite{hendrycks2022scaling}, $k$\textsuperscript{th} Nearest Neighbor (KNN) \cite{sun2022out}, Virtual-Logit Matching (ViM) \cite{wang2022vim}, and Generalized Entropy (GEN) \cite{liu2023gen}. They are chosen on the basis of popularity, competitiveness, and diversity in principle, in which information are leveraged from the feature, logit, and/or softmax probability spaces to handcraft a granular OOD scoring function beyond conventional uncertainty measures. Note, some methods are more prohibitive from needing access to ID training samples, which may be confidential, or embedding that can be inaccessible in black-box models; both are pertinent in the context of DP.

\begin{table*}[t]
\small
\centering
\caption{\textbf{Overview of frequentist methods.} Feat. refers to the classifier's penultimate activation map. We follow the original paper's guidelines to set hyperparameters where $c$ and $d$ is the class number and the feat's dimension.}
\resizebox{0.8\textwidth}{!}{%
\begin{tabular}{lccll}
\toprule
Detector & Space & ID data free? & Working principle & Hyperparameter  \\
\midrule
MSP \scriptsize{(ICLR'17 \cite{hendrycks2016baseline})} & Prob. & Y & Max softmax prob. & \textit{n/a} \\
Maha \scriptsize{(NeurIPS'18 \cite{lee2018simple})} & Feat. & N & Min feat. dist. to train set's classwise centroids & \textit{n/a} \\
R+E \scriptsize{(NeurIPS'21 \cite{sun2021react})} & Feat./Logit & N & Feat. truncation followed by energy function & Truncate by ID's 98\%. \\
GrN \scriptsize{(NeurIPS'21 \cite{huang2021importance})} & Prob. & Y & Backpropagated gradients & \textit{n/a} \\
MLS \scriptsize{(ICML'21 \cite{hendrycks2022scaling})} & Logit & Y & Max logit & \textit{n/a} \\
KLM \scriptsize{(ICML'21 \cite{hendrycks2022scaling})} & Prob. & N & Min KL div. \textit{w.r.t.} the train set's class-wise probs. & \textit{n/a} \\
KNN \scriptsize{(ICML'22 \cite{sun2022out})} & Feat. & N & $k$-th nearest neighbor dist. \textit{w.r.t.} train set's feats. & $k\approx2.5\times c$ \\
ViM \scriptsize{(CVPR'22 \cite{wang2022vim})} & Feat./Logit & N & Norm of feat's residual projection, plus energy & Subspace dim. $\approx d/2$ \\
GEN \scriptsize{(CVPR'23 \cite{liu2023gen})} & Prob. & Y & Generalized entropy of prob. & $\gamma = 0.1$; no suppression \\
\bottomrule
\end{tabular}}
\label{tab2}
\end{table*}

As hinted, \textbf{UQ} orthogonally provides a means to detect OOD as well. Hence, we inspect Deep Ensembles (DE) \cite{lakshminarayanan2017simple}, the UQ gold-standard, using a heterogeneous setup comprised of four members and evaluate two uncertainty measures: Total Uncertainty (TU), the Shannon entropy of the DE-averaged probability, and Epistemic Uncertainty (EU), the mutual information across the members' probabilities. Unlike the post-hoc schemes, DE incurs overhead from training multiple models and performing multiple forward-passes during inference. 

\textbf{Metrics.} For S-OODD, we report the AUROC ($\uparrow$; arrow indicates which direction is better). For MC-OODD, we report the PPR ($\uparrow$) \cite{malinin2019ensemble}, where closer to 100\% indicates correlation of low confidence to mispredictions; conversely, zero/negative values mean no/anti-correlation and is thus undesirable. We refer to \cite{malinin2019ensemble} for full calculation details. We also report the class-balanced accuracy (Acc., $\uparrow$) of the ID test and covariate OOD sets to present a full picture on robustness (\textit{e.g.}, a model with trivial performance can display near zero accuracy but high PRR). Note, when using the DE as the detector, we report the accuracy over the DE-averaged predictions.

\begin{table*}[!b]
\small
\centering
\caption{\textbf{(C1)} Results of various TL configs in ResNet-50.}
\resizebox{0.72\textwidth}{!}{%
\begin{tabular}{l||c|ccccccccc||c|cc}
\toprule
& & \multicolumn{8}{c}{Single-Model} & & \multicolumn{3}{c}{DE \scriptsize{\cite{lakshminarayanan2017simple}} (×4)} \\
Pre-train (TL) & Acc\% & MSP \scriptsize{\cite{hendrycks2016baseline}} & Maha \scriptsize{\cite{lee2018simple}} & R+E \scriptsize{\cite{sun2021react}} & GrN \scriptsize{\cite{huang2021importance}} & MLS \scriptsize{\cite{hendrycks2022scaling}} & KLM \scriptsize{\cite{hendrycks2022scaling}} & KNN \scriptsize{\cite{sun2022out}} & ViM \scriptsize{\cite{wang2022vim}} & GEN \scriptsize{\cite{liu2023gen}} & Acc\% & TU & EU \\
\midrule
\midrule 

\multicolumn{14}{l}{\textbf{\circled{S1} BreakHis: ID Acc\% \& S-OODD AUROC\%}} \\
None & 73.31 & 57.85 & 61.58 &  57.39 & 54.11 & 56.89 & 58.4 & \cellcolor{blue!25}70.15 & \cellcolor{orange!25}62.33 & 56.47 & 74.04 & 59.24 & \cellcolor{green!25}61.83 \\
IN1K & 43.82 & 64.86 & 48.94 & 62.14 & 64.63 & \cellcolor{green!25}65.62 & 53.58 & 64.14 & 48.60 & \cellcolor{orange!25}65.68 & 47.75 & \cellcolor{blue!25}67.19 & 49.63 \\
MoCo v2 \scriptsize{\cite{chen2020improved, kang2023benchmarking}} & 95.42 & 72.22 & \cellcolor{blue!25}82.53 & 74.40 & 73.15 & 73.69 & 70.72 & 74.62 & \cellcolor{orange!25}81.94 & 75.31 & 95.88 & 77.30 & \cellcolor{green!25}77.60 \\
SwAV \scriptsize{\cite{caron2020unsupervised, kang2023benchmarking}} & \textbf{96.61} & \textbf{79.40} & \cellcolor{green!25}\textbf{83.44} & \textbf{82.01} & \textbf{78.16} & \textbf{81.99} & \textbf{76.74} & \textbf{75.35} & \cellcolor{blue!25}\textbf{{86.37}} & \textbf{82.95} & \textbf{97.02} & \cellcolor{orange!25}\textbf{83.86} & 77.09 \\
BT \scriptsize{\cite{zbontar2021barlow, kang2023benchmarking}} & 94.89 & 72.81 & \cellcolor{orange!25}80.79 & 76.25 & 66.83 & 76.22 & 73.09 & 74.23 & \cellcolor{blue!25}81.05 & 75.98 & 95.73 & 76.89 & \cellcolor{green!25}\textbf{78.76} \\
\midrule

\multicolumn{14}{l}{\textbf{\circled{S2} NCT-CRC: ID Acc\% \& S-OODD AUROC\%}} \\
None & 78.92 & 56.64 & 59.14 & 59.59 & 55.29 & 56.86 & \cellcolor{orange!25}66.33 & \cellcolor{blue!25}69.01 & \cellcolor{green!25}64.92 & 56.96 & 79.96 & 57.84 & 62.38 \\
IN1K & 93.83 & \textbf{84.72} & 76.45 & \cellcolor{green!25}84.99 & 70.77 & \textbf{84.86} & \textbf{83.86} & \cellcolor{orange!25}87.25 & 84.38 & 83.16 & 94.62 & \cellcolor{blue!25}\textbf{87.47} & 76.31 \\
MoCo v2 \scriptsize{\cite{chen2020improved, kang2023benchmarking}} & 95.19 & 78.37 & \cellcolor{green!25}82.61 & 78.88 & \textbf{79.87} & 78.57 & 76.59 & 79.88 & \cellcolor{blue!25}85.20 & 79.27 & 95.69 & \cellcolor{orange!25}83.07 & 81.72 \\
SwAV \scriptsize{\cite{caron2020unsupervised, kang2023benchmarking}} & 95.78 & 80.86 & \cellcolor{orange!25}87.52 & 86.39 & 72.96 & 82.43 & 82.08 & \cellcolor{green!25}86.98 & \cellcolor{blue!25}88.59 & 82.80 & 96.28 & 84.44 & 78.62 \\
BT \scriptsize{\cite{zbontar2021barlow, kang2023benchmarking}} & \textbf{96.18} & 81.82 & \cellcolor{green!25}\textbf{87.79} & \textbf{86.64} & 77.06 & 84.80 & 82.86 & \cellcolor{blue!25}\textbf{{92.08}} & \cellcolor{orange!25}\textbf{91.44} & \textbf{85.21} & \textbf{96.81} & 87.30 & \textbf{84.19} \\
\midrule

\multicolumn{14}{l}{\textbf{\circled{S3} BreakHis: Covariate OOD Acc\% \& MC-OODD PRR\%}} \\
None & 45.64 & \cellcolor{blue!25}18.84 & 1.33 & 11.67 & 11.29 & 12.54 & 9.22 & 9.41 & 3.04 & 11.37 & 46.11 & \cellcolor{orange!25}17.29 & \cellcolor{green!25}12.80 \\
IN1K & 28.91 & \cellcolor{green!25}21.53 & -8.61 & 12.55 & 19.51 & \cellcolor{orange!25}21.59 & 4.02 & 16.17 & -7.36 & 20.72 & 30.47 & \cellcolor{blue!25}22.43 & -9.55 \\
MoCo v2 \scriptsize{\cite{chen2020improved, kang2023benchmarking}} & \textbf{58.41} & 21.12 & \cellcolor{orange!25}33.34 & 16.23 & 17.88 & 18.26 & 21.78 & 22.40 & \cellcolor{blue!25}34.44 & 18.81 & \textbf{59.91} & 25.39 & \cellcolor{green!25}30.16 \\
SwAV \scriptsize{\cite{caron2020unsupervised, kang2023benchmarking}} & 57.30 & \textbf{34.05} & 33.41 & \textbf{31.41} & \textbf{34.19} & \textbf{33.04} & \cellcolor{green!25}\textbf{34.81} & 33.66 & \cellcolor{orange!25}\textbf{36.56} & \textbf{34.30} & 57.95 & \cellcolor{blue!25}\textbf{{37.41}} & 19.38 \\
BT \scriptsize{\cite{zbontar2021barlow, kang2023benchmarking}} & 58.23 & 27.79 & \textbf{34.66} & 27.62 & 28.27 & 27.18 & 27.25 & \cellcolor{blue!25}\textbf{37.14} & \cellcolor{orange!25}35.55 & 27.68 & 59.43 & 34.11 & \cellcolor{green!25}\textbf{35.47} \\
\midrule

\multicolumn{14}{l}{\textbf{\circled{S4} NCT-CRC: Covariate OOD Acc\% \& MC-OODD PRR\%}} \\
None & 53.06 & 15.85 & 11.34 & 7.25 & 5.24 & 7.44 & \cellcolor{green!25}16.00 & \cellcolor{blue!25}25.00 &	13.32 & 7.46 & 53.30 & 15.88 & \cellcolor{orange!25}17.12 \\
IN1K & 73.47 & \cellcolor{blue!25}50.40 & 29.11 & 45.10 & 36.47 & \cellcolor{green!25}48.76 & 47.58 & 46.66 & 36.51 & 45.18 & 75.07 & \cellcolor{orange!25}50.37 & 25.58 \\
MoCo v2 \scriptsize{\cite{chen2020improved, kang2023benchmarking}} & 80.19 & \cellcolor{green!25}49.29 & 42.60 & 46.87 & \cellcolor{orange!25}\textbf{49.32} & 48.11 & 45.95 & 48.84 & 48.11 & 48.18 & 82.32 & \cellcolor{blue!25}53.22 & 49.26 \\
SwAV \scriptsize{\cite{caron2020unsupervised, kang2023benchmarking}} & \textbf{81.75} & \cellcolor{green!25}\textbf{61.68} & \textbf{51.62} & \textbf{59.34} & 48.45 & \textbf{60.18} & \cellcolor{orange!25}\textbf{61.80} & \textbf{60.72} & 53.20 & \textbf{58.35} & 83.31 & \cellcolor{blue!25}\textbf{{62.62}} & 50.42 \\
BT \scriptsize{\cite{zbontar2021barlow, kang2023benchmarking}} & 81.63 & \cellcolor{orange!25}58.01 & 49.36 & 52.80 & 40.30 & 53.95 & \cellcolor{green!25}55.97 & 55.47 & \textbf{54.46} & 53.83 & \textbf{83.56} & \cellcolor{blue!25}62.00 & \textbf{55.19} \\

\bottomrule
\end{tabular}}
\label{tab3}
\end{table*}

\begin{table*}[!b]
\small
\centering
\caption{\textbf{(C2)} Results of various TL configs in ViT-B/16.}
\resizebox{0.72\textwidth}{!}{%
\begin{tabular}{l||c|ccccccccc||c|cc}
\toprule
& & \multicolumn{8}{c}{Single-Model} & & \multicolumn{3}{c}{DE \scriptsize{\cite{lakshminarayanan2017simple}} (×4)} \\
Pre-train (TL) & Acc\% & MSP \scriptsize{\cite{hendrycks2016baseline}} & Maha \scriptsize{\cite{lee2018simple}} & R+E \scriptsize{\cite{sun2021react}} & GrN \scriptsize{\cite{huang2021importance}} & MLS \scriptsize{\cite{hendrycks2022scaling}} & KLM \scriptsize{\cite{hendrycks2022scaling}} & KNN \scriptsize{\cite{sun2022out}} & ViM \scriptsize{\cite{wang2022vim}} & GEN \scriptsize{\cite{liu2023gen}} & Acc\% & TU & EU \\
\midrule
\midrule 

\multicolumn{14}{l}{\textbf{\circled{S1} BreakHis: ID Acc\% \& S-OODD AUROC\%}} \\
None & 43.28 & 53.26 & \cellcolor{green!25}58.92 & 54.81 & 53.89 & 54.01 & 55.33 & \cellcolor{blue!25}63.37 & \cellcolor{orange!25}60.49 & 54.70 & 47.75 & 54.47 & 57.87 \\
IN1K & 95.06 & 68.58 & \cellcolor{orange!25}77.43 & 73.85 & 67.27 & 73.70 & 68.26 & 72.22 & \cellcolor{blue!25}78.98 & 73.57 & \textbf{96.60} & 75.06 & \cellcolor{green!25}76.04 \\
CLIP \scriptsize{\cite{radford2021learning}} & 56.93 & 55.96 & \cellcolor{orange!25}63.97 & 59.35 & 56.33 & 58.57 & 54.80 & \cellcolor{green!25}60.90 & \cellcolor{blue!25}65.09 & 58.63 & 69.45 & 58.88 & 54.18 \\
BiomedCLIP \scriptsize{\cite{zhang2023large}} & \textbf{95.49} & \textbf{73.71} & \textbf{77.74} & \textbf{78.25} & \textbf{74.42} & \textbf{78.51} & \textbf{74.16} & \textbf{76.18} & \cellcolor{blue!25}\textbf{{80.81}} & \textbf{78.51} & \textbf{96.60} & \cellcolor{green!25}\textbf{79.17} & \cellcolor{orange!25}\textbf{80.68} \\
QuiltNet \scriptsize{\cite{ikezogwo2024quilt}} & 60.16 & 56.03 & \cellcolor{orange!25}64.17 & 60.25 & 59.40 & 59.36 & 54.02 & \cellcolor{green!25}62.08 & \cellcolor{blue!25}66.13 & 60.07 & 67.51 & 61.19 & 60.12 \\
\midrule

\multicolumn{14}{l}{\textbf{\circled{S2} NCT-CRC: ID Acc\% \& S-OODD AUROC\%}} \\
None & 76.85 & 63.63 & \cellcolor{green!25}67.55 & 62.24 & 64.09 & 63.20 & 66.95 & \cellcolor{orange!25}68.35 & \cellcolor{blue!25}69.98 & 63.50 & 78.00 & 66.36 & 65.83 \\
IN1K & \textbf{97.21} & \textbf{86.03} & \textbf{84.20} & \cellcolor{green!25}\textbf{87.86} & 82.30 & \textbf{87.46} & \textbf{86.62} & 86.21 & \textbf{87.24} & \cellcolor{orange!25}\textbf{87.98} & \textbf{97.84} & \cellcolor{blue!25}\textbf{{90.73}} & \textbf{87.38} \\
CLIP \scriptsize{\cite{radford2021learning}} & 93.91 & 71.26 & \cellcolor{green!25}78.77 & 72.76 & 72.93 & 73.02 & 74.48 & \cellcolor{blue!25}81.95 & \cellcolor{orange!25}79.84 & 73.03 & 96.33 & 75.61 & 73.2 \\
BiomedCLIP \scriptsize{\cite{zhang2023large}} & 96.49 & 83.95 & 74.23 & \cellcolor{green!25}86.78 & \textbf{82.94} & 86.52 & 84.99 & \textbf{86.48} & 83.36 & \cellcolor{orange!25}86.89 & 97.04 & \cellcolor{blue!25}87.92 & 85.76 \\
QuiltNet \scriptsize{\cite{ikezogwo2024quilt}} & 95.02 & 74.38 & \cellcolor{green!25}81.73 & 72.05 & 73.43 & 73.09 & 75.75 & \cellcolor{blue!25}83.18 & \cellcolor{orange!25}82.29 & 73.94 & 96.44 & 78.79 & 77.80 \\
\midrule

\multicolumn{14}{l}{\textbf{\circled{S3} BreakHis: Covariate OOD Acc\% \& MC-OODD PRR\%}} \\
None & 37.06 & \cellcolor{blue!25}20.50 & 6.05 & 14.46 & 15.70 & \cellcolor{green!25}17.04 & 2.52 & 11.93 & 10.27 & 14.84 & 39.65 & \cellcolor{orange!25}20.18 & 9.73 \\
IN1K & 67.37 & \cellcolor{orange!25}41.77 & 18.16 & 31.31 & 32.80 & 34.48 & 34.96 & 22.51 & 21.49 & \cellcolor{green!25}35.74 & 70.47 & \cellcolor{blue!25}45.05 & 33.34 \\
CLIP \scriptsize{\cite{radford2021learning}} & 41.19 & \cellcolor{orange!25}19.37 & 4.92 & 16.56 & 15.87 & \cellcolor{green!25}18.27 & 2.43 & 7.31 & 7.74 & 17.23 & 48.46 & \cellcolor{blue!25}20.65 & 8.88 \\
BiomedCLIP \scriptsize{\cite{zhang2023large}} & \textbf{70.07} & \cellcolor{orange!25}\textbf{48.61} & \textbf{26.45} & \textbf{39.81} & \textbf{42.83} & \textbf{43.14} & \cellcolor{green!25}\textbf{45.60} & \textbf{34.71} & \textbf{32.95} & \textbf{44.09} & \textbf{73.14} & \cellcolor{blue!25}\textbf{{52.83}} & \textbf{42.77} \\
QuiltNet \scriptsize{\cite{ikezogwo2024quilt}} & 44.06 & \cellcolor{orange!25}20.69 & 5.26 & 17.50 & 17.64 & \cellcolor{green!25}19.98 & 1.30 & 11.42 & 9.74 & 18.39 & 48.89 & \cellcolor{blue!25}20.81 & 9.96 \\
\midrule

\multicolumn{14}{l}{\textbf{\circled{S4} NCT-CRC: Covariate OOD Acc\% \& MC-OODD PRR\%}} \\
None & 63.87 & \cellcolor{blue!25}40.89 & 8.10 & 16.19 & \cellcolor{green!25}33.64 & 23.30 & 29.48 & 24.13 & 12.23 & 22.08 & 65.04 & \cellcolor{orange!25}37.35 & 25.27 \\
IN1K & \textbf{90.13} & \cellcolor{blue!25}\textbf{{75.09}} & \textbf{46.39} & \textbf{65.11} & \textbf{67.61} & \textbf{69.29} & \cellcolor{green!25}\textbf{72.90} & \textbf{57.45} & \textbf{51.61} & \textbf{68.66} & \textbf{91.70} & \cellcolor{orange!25}\textbf{74.92} & \textbf{62.14} \\
CLIP \scriptsize{\cite{radford2021learning}} & 75.15 & \cellcolor{orange!25}45.88 & 33.14 & 36.35 & 40.58 & 40.48 & 41.99 & \cellcolor{green!25}42.18 & 35.99 & 40.05 & 78.29 & \cellcolor{blue!25}49.63 & 36.70 \\
BiomedCLIP \scriptsize{\cite{zhang2023large}} & 87.03 & \cellcolor{blue!25}68.05 & 27.74 & 58.42 & 62.38 & 62.00 & \cellcolor{green!25}65.39 & 56.36 & 43.15 & 62.14 & 88.76 & \cellcolor{orange!25}68.03 & 58.38 \\
QuiltNet \scriptsize{\cite{ikezogwo2024quilt}} & 77.17 & \cellcolor{orange!25}49.40 & 43.37 & 39.91 & 42.37 & 44.75 & 46.18 & \cellcolor{green!25}48.04 & 44.20 &	43.68 & 79.99 & \cellcolor{blue!25}53.54 & 46.49 \\

\bottomrule
\end{tabular}}
\label{tab4}
\end{table*}

\section{Results and Discussion}
We report the results of experimental configs. (C1), (C2), (C3) in Tabs. \ref{tab3}-\ref{tab5}, respectively. For clarity, in each table, we organize our results into four row \textit{Sections}: \circled{S1} and \circled{S2} lists the \textit{ID Acc.} and \textit{each detector's S-OODD AUROC} over BreakHis and NCT-CRC, respectively. Note, we display ID Acc. separately for DE. We report the averaged metrics over the entire OSR splits, each with 10 trials using different random seeds. Similarly, \circled{S3} and \circled{S4} respectively shows the \textit{covariate OOD Acc.} and \textit{each detector's MC-OODD PRR} over BreakHis and NCT-CRC. The best metric across the TL (Tabs. \ref{tab3}-\ref{tab4}) or architecture (Tab. \ref{tab5}) configs are in \textbf{bold} (\textit{column-wise direction}). For each TL/architecture config, we further highlight the top-3 (\hlc[blue!25]{1\textsuperscript{st}}, \hlc[orange!25]{2\textsuperscript{nd}}, \hlc[green!25]{3\textsuperscript{rd}}) performant detectors (\textit{row-wise}).

\textbf{Does DP domains pose more challenges?} Not really -- At the very least, the AUROC and PRR of our experiments do not deviate from the statistics therein natural imagery benchmarks \cite{zhang2023openood, pinto2022impartial}. This is promising as it suggests that \textit{we can tackle this problem in a general framework catered to natural images}. However, another popular line of research is Outlier Exposure (OE), in which an auxiliary OOD set is exploited during training to encourage low confidence. Unlike natural imagery where a plethora of effective OE candidate sets are publicly available, collecting or synthesizing such a dataset in DP is challenging, even unfeasible. We leave this to future work.

\begin{table*}[!t]
\small
\centering
\caption{\textbf{(C3)} Results of diff. architectures using our SIAYN TL. We chose size variants of ConvNeXt, ViT, Swin to fit in 24GB of GPU.}
\resizebox{0.72\textwidth}{!}{%
\begin{tabular}{l||c|ccccccccc||c|cc}
\toprule
& & \multicolumn{8}{c}{Single-Model} & & \multicolumn{3}{c}{DE \scriptsize{\cite{lakshminarayanan2017simple}} (×4)} \\
Pre-train (TL) & Acc\% & MSP \scriptsize{\cite{hendrycks2016baseline}} & Maha \scriptsize{\cite{lee2018simple}} & R+E \scriptsize{\cite{sun2021react}} & GrN \scriptsize{\cite{huang2021importance}} & MLS \scriptsize{\cite{hendrycks2022scaling}} & KLM \scriptsize{\cite{hendrycks2022scaling}} & KNN \scriptsize{\cite{sun2022out}} & ViM \scriptsize{\cite{wang2022vim}} & GEN \scriptsize{\cite{liu2023gen}} & Acc\% & TU & EU \\
\midrule
\midrule 

\multicolumn{14}{l}{\textbf{\circled{S1} BreakHis: ID Acc\% \& S-OODD AUROC\%}} \\
ResNet-50 & 93.98 & 76.25 & 79.15 & 79.11 & 74.41 & 79.04 & 73.43 & 74.47 & \cellcolor{blue!25}83.33 & \cellcolor{green!25}79.29 & 94.43 & \cellcolor{orange!25}80.22 & 73.70 \\
ConvNeXt-S \scriptsize{\cite{liu2022convnet}} & \textbf{96.47} & \textbf{78.38} & 83.64 & \textbf{81.71} & \textbf{78.31} & \textbf{82.02} & \textbf{78.43} & 74.63 & 85.13 & \cellcolor{green!25}\textbf{82.60} & 97.40 & \cellcolor{orange!25}\textbf{82.75} & \cellcolor{blue!25}\textbf{82.84} \\
ViT-B/16 \scriptsize{\cite{dosovitskiy2020image}} & 96.22 & 77.20 & 80.50 & 80.59 & 71.71 & 80.62 & 74.86 & 72.66 & \cellcolor{blue!25}82.13 & \cellcolor{green!25}80.80 & 96.94 & \cellcolor{orange!25}81.55 & 80.62 \\
Swin-T \scriptsize{\cite{liu2021swin}} & 96.37 & 74.75 & \cellcolor{orange!25}\textbf{85.28} & 79.35 & 71.48 & 79.27 & 74.27 & \textbf{75.40} & \cellcolor{blue!25}\textbf{{87.05}} & 79.71 & \textbf{97.41} & 80.07 & \cellcolor{green!25}81.10 \\
\midrule

\multicolumn{14}{l}{\textbf{\circled{S2} NCT-CRC: ID Acc\% \& S-OODD AUROC\%}} \\
ResNet-50 & 95.37 & 82.51 & 78.08 & \cellcolor{orange!25}84.49 & 72.92 & 83.78 & 82.20 & \cellcolor{green!25}84.03 & 82.68 & 82.47 & 95.95 & \cellcolor{blue!25}85.04 & 79.45 \\
ConvNeXt-S \scriptsize{\cite{liu2022convnet}} & 96.04 & \textbf{86.20} & \cellcolor{orange!25}\textbf{{92.76}} & \textbf{87.65} & \textbf{86.03} & \textbf{87.52} & 85.72 & \cellcolor{green!25}\textbf{92.74} & \cellcolor{blue!25}\textbf{92.84} & \textbf{88.38} & 97.06 & \textbf{89.42} & 85.94 \\
ViT-B/16 \scriptsize{\cite{dosovitskiy2020image}} & \textbf{96.79} & 85.64 & \cellcolor{orange!25}92.08 & 86.73 & 80.07 & 86.58 & \textbf{86.69} & \cellcolor{green!25}90.79 & \cellcolor{blue!25}92.69 & 86.96 & \textbf{97.62} & 88.77 & \textbf{88.09} \\
Swin-T \scriptsize{\cite{liu2021swin}} & 95.82 & 82.07 & \cellcolor{orange!25}91.45 & 84.96 & 77.60 & 84.57 & 81.91 & \cellcolor{green!25}89.48 & \cellcolor{blue!25}91.95 & 84.89 & 96.77 & 86.56 & 86.20 \\
\midrule

\multicolumn{14}{l}{\textbf{\circled{S3} BreakHis: Covariate OOD Acc\% \& MC-OODD PRR\%}} \\
ResNet-50 & 53.06 & 26.66 & 17.44 & \cellcolor{green!25}27.03 & 17.91 & 26.09 & \cellcolor{blue!25}28.90 & 21.83 & 19.70 & 22.74 & 54.34 & \cellcolor{orange!25}27.78 & 17.41 \\
ConvNeXt-S \scriptsize{\cite{liu2022convnet}} & 62.93 & \cellcolor{orange!25}\textbf{45.78} & \textbf{26.82} & \textbf{36.68} & \textbf{42.33} & \textbf{41.40} & \cellcolor{green!25}\textbf{42.74} & \textbf{31.63} & \textbf{29.86} & \textbf{41.04} & 65.63 & \cellcolor{blue!25}47.82 & \textbf{37.10} \\
ViT-B/16 \scriptsize{\cite{dosovitskiy2020image}} & \textbf{67.37} & \cellcolor{orange!25}45.71 & 9.80 & 36.35 & 37.92 & 40.19 & \cellcolor{green!25}42.64 & 19.27 & 17.27 & 40.62 & \textbf{69.61} & \cellcolor{blue!25}\textbf{{47.98}} & 35.95 \\
Swin-T \scriptsize{\cite{liu2021swin}} & 62.07 & \cellcolor{orange!25}39.72 & 23.49 & 32.02 & 33.92 & 35.32 & \cellcolor{green!25}36.68 & 27.38 & 26.82 & 35.79 & 64.38 & \cellcolor{blue!25}42.76 & 35.28 \\
\midrule

\multicolumn{14}{l}{\textbf{\circled{S4} NCT-CRC: Covariate OOD Acc\% \& MC-OODD PRR\%}} \\
ResNet-50 & 76.52 & \cellcolor{blue!25}49.29 & 34.53 & 43.90 & 34.08 & \cellcolor{green!25}48.27 & 47.88 & 41.83 & 37.49 & 42.14 & 78.81 & \cellcolor{orange!25}48.99 & 34.51 \\
ConvNeXt-S \scriptsize{\cite{liu2022convnet}} & \textbf{84.88} & \cellcolor{blue!25}\textbf{{63.53}} & \textbf{37.85} & \textbf{52.21} & \textbf{58.36} & \textbf{57.85} & \cellcolor{green!25}\textbf{62.03} & \textbf{51.01} & \textbf{40.92} & \textbf{56.64} & \textbf{88.47} & \cellcolor{orange!25}\textbf{62.19} & 52.07 \\
ViT-B/16 \scriptsize{\cite{dosovitskiy2020image}} & 84.38 & \cellcolor{blue!25}61.44 & 23.11 & 50.55 & 55.56 & 54.76 & \cellcolor{green!25}59.21 & 43.61 & 30.65 & 54.52 & 87.77 & \cellcolor{orange!25}59.45 & \textbf{52.22} \\
Swin-T \scriptsize{\cite{liu2021swin}} & 79.30 & \cellcolor{orange!25}54.44 & 35.40 & 47.21 & 48.57 & 50.44 & \cellcolor{green!25}51.49 & 46.31 & 37.76 & 50.81 & 83.09 & \cellcolor{blue!25}56.70 & 50.85 \\

\bottomrule
\end{tabular}}
\label{tab5}
\end{table*}

\textbf{Accuracy can be deceptive.} While accuracy generally translates to S-OODD and MC-OODD performance, solely relying on it to rank TL or architecture configs. when all accuracies are within $\pm$1\% of each other is precarious. For instance, ViT-B/16 in Tab. \ref{tab5} displays the highest accuracy in \circled{S2} \& \circled{S3} but seldom beats ConvNeXt-S in S-OODD or MC-OODD across most detectors. 

\textbf{No universal SOTA detector.} The ranking among detectors in Tabs. \ref{tab3}-\ref{tab5} is volatile, sensitive to the dataset, TL, and architecture. In general though, \textbf{ViM is superior in S-OODD} followed by KNN, Maha, TU, and GEN. Conversely, \textbf{TU and MSP excels in MC-OODD} followed by KLM and ViM. As a rule of thumb, we recommend these detectors as an initial choice. From these trends, we surmise that the information from the rich feature representation is vital for S-OODD, whereas MC-OODD benefits more when the detector operates near the final output space. \textit{Thus, achieving all-round superior performance in both tasks via a single detector is elusive}. Note, the differences in practical constraint (\textit{e.g.}, ViM, KNN, Maha require ID samples and the feature embedding, while TU adopts the costly DE framework, which is not suited for rapid or mobile applications) further underscores this ``\textit{no winner}'' posture.

\textbf{The contradiction of UQ.} Uncertainty due to OOD-ness must \textit{axiomatically} be encoded in EU, whereas TU conflates OOD and aleatoric sources (AU, \textit{e.g.}, ambiguity near class boundaries). Surprisingly though, \textit{EU detectors often underperform TU}, further supported by \cite{ulmer2021know, wimmer2023quantifying, henning2021bayesian}. Is this a problem? That is, can't we just use TU regardless of such an axiomatic contradiction? While doing so imposed no practical setback in our experiments, we anticipate it to be a problem when fuzzy data are non-negligible or the disentanglement of AU-EU sources is crucial (\textit{e.g.}, in active learning, we generally wish to avoid querying fuzzy data). 

\textbf{Is DE worth the extra cost?} The combination of ViM+MSP enjoys comparable/superior performance to the DE. \textit{Hence, a powerful frequentist detector(s) may be all you need}. This is promising news for efficient DP as novelties and errors can be sought in a seamless and timely fashion. We recognize though that ensembling always boosts accuracy, and thus, DE holds merit in this respect.

\textbf{DP-based TL helps, albeit unpredictably.} We see compelling improvement over de-novo. However, further statements cannot be made. For instance, there is no clear winner among the SSL methods (Tab. \ref{tab3}), \textit{e.g.}, SwAV is superior in S-OODD over BreakHis but never in NCT-CRC. QuiltNet also surprisingly fails to outperform BiomedCLIP (Tab. \ref{tab4}), in spite of the exposure to H\&E in the former but not the latter; in fact, Quiltnet's gain is relatively marginal in BreakHis. Moreover, SIAYN (Tab. \ref{tab5}) is unexpectedly effective, sometimes on par or surpassing their SSL and LVLM counterparts (\textit{e.g.}, MoCo v2, BiomedCLIP), despite its much smaller scale. \textit{Although we affirm the advantages of DP TL, these examples (some counterintuitive) demonstrate that much research is still needed to codify its exact effects}.

\textbf{What about TL with natural images?} Such TL still often helps, sometimes more than DP TL as demonstrated by the NCT-CRC experiments in Tab. \ref{tab4} wherein IN1K outperformed BiomedCLIP. In large, however, it is less advantageous, even performing worse than de-novo in certain cases (BreakHis experiments in Tab. \ref{tab3}). Our verdict is, \textit{TL with natural imagery on average helps, however, further research is likewise needed (especially on when it fails) and should be implemented with caution}. TL from DP is the safer option and we recommend IN1K only when it is unavailable. In this spirit, we extend our SIAYN checkpoints to over ten popular, off-the-shelf DNN architectures and release them with our code.

\textbf{ConvNeXt is robust, as are transformers.} Recent studies to date have been dedicated to explaining the superior robustness properties of transformers \cite{bai2021transformers}. However, \cite{pinto2022impartial} showed that more advanced CNNs like ConvNeXt can behave just as robustly. Looking at Tab. \ref{tab5}, our findings agree with the latter study wherein ConvNeXt-S surpass ViT-B/16 and Swin-T in more number of metrics. Nonetheless, transformers perform well too and we conclude by saying \textit{all three are SOTA}. However, not all CNNs are equal as we see a considerable gap between ResNet-50 and ConvNeXt-S. Hence, lumping them as one or analyzing just the ResNet family of architectures may give a false impression of CNNs. From this result, \textit{we also urge to move away from ResNets, which remain a popular choice in DP, and onto more SOTA CNNs like ConvNeXt}.

\section{Concluding Remark}
We present an in-depth robustness study on S-OODD and MC-OODD in DP, with an emphasis on proper procedures and diverse settings. We reveal insights, some of which challenge the status quo taken for granted, \textit{e.g.}, DE uncertainties may be redundant, TL from DP may not always behave as expected, CNNs can be just as robust as transformers. We hope our findings serve as a stepping stone to further exploration of pertinent themes in the realms of DP.

\begin{credits}
\subsubsection{\ackname} Research reported in this publication was supported by the National Institutes of Health under award numbers R01EB009745, R01CA260830, and R21CA263147. The content is solely the responsibility of the authors and does not necessarily represent the official views of the National Institutes of Health. This work was also supported by NSF grant 2243257, the National Science Foundation Science and Technology Center for Quantitative Cell Biology. Rohit Bhargava is a CZ Biohub Investigator.
 
\subsubsection{\discintname}
The authors declare no conflict of interest.
\end{credits}

\bibliographystyle{splncs04}
\bibliography{bib}

\end{document}